\documentclass[sigconf,authorversion]{acmart}

\renewcommand\footnotetextcopyrightpermission[1]{}

\usepackage{amsmath}

\usepackage{amssymb}
\usepackage{booktabs}
\usepackage{graphicx}
\graphicspath{{figures/}}
\usepackage{microtype}
\usepackage{multirow}
\usepackage{enumitem}
\usepackage{xcolor}
\usepackage{placeins}
\usepackage{float}
\usepackage{tikz}
\usetikzlibrary{arrows.meta,positioning,fit,backgrounds,calc}

\setlist{nosep,leftmargin=1.2em}

\copyrightyear{2026}
\setcopyright{none}
\acmConference{}{}{}
\acmBooktitle{}
\acmDOI{}
\acmISBN{}

\begin{document}

\title{SCNO: Spiking Compositional Neural Operator --- Towards a Neuromorphic Foundation Model for Nuclear PDE Solving}

\author{Samrendra Roy}
\email{roysam@illinois.edu}
\affiliation{%
 \institution{University of Illinois Urbana-Champaign}
 \department{Department of Nuclear, Plasma \& Radiological Engineering}
 \city{Urbana}
 \state{IL}
 \country{USA}
}

\author{Souvik Chakraborty}
\affiliation{%
 \institution{Indian Institute of Technology Delhi}
 \department{Department of Applied Mechanics}
 \city{New Delhi}
 \country{India}
}

\author{Rizwan-uddin}
\affiliation{%
 \institution{University of Illinois Urbana-Champaign}
 \department{Department of Nuclear, Plasma \& Radiological Engineering}
 \city{Urbana}
 \state{IL}
 \country{USA}
}

\author{Syed Bahauddin Alam}
\affiliation{%
 \institution{University of Illinois Urbana-Champaign}
 \department{Department of Nuclear, Plasma \& Radiological Engineering}
 \city{Urbana}
 \state{IL}
 \country{USA}
}
\affiliation{%
 \institution{National Center for Supercomputing Applications}
 \city{Urbana}
 \state{IL}
 \country{USA}
}

\ccsdesc[500]{Hardware~Neural systems}
\ccsdesc[500]{Computing methodologies~Neural networks}
\ccsdesc[300]{Computing methodologies~Transfer learning}

\begin{abstract}
Neural operators have emerged as powerful surrogates for partial differential equation (PDE) solvers, yet they are typically trained as monolithic models for individual PDEs, require energy-intensive GPU hardware, and must be retrained from scratch when new physics emerge.
We introduce the \emph{Spiking Compositional Neural Operator} (SCNO), a modular architecture combining spiking and conventional components that addresses all three limitations.
SCNO maintains a library of small spiking neural operator blocks, each trained on a single elementary differential operator (convection, diffusion, reaction), and composes them through a lightweight input-conditioned aggregator to solve coupled PDEs not seen during block training.
A small correction network learns cross-coupling residuals while keeping all blocks and the aggregator frozen, preserving zero-forgetting modular expansion by construction.
We evaluate SCNO on eight PDE families including five coupled systems and a nuclear-relevant 1-group neutron diffusion equation.
SCNO with correction achieves the lowest relative $L^2$ error on four of five coupled PDEs, outperforming both a monolithic spiking DeepONet (by up to 62\%, mean over 3 seeds) and a standard ANN DeepONet (by up to 65\%), while requiring only 95K trainable parameters versus 462K for the monolithic baseline.
To our knowledge, this is the first compositional spiking neural operator and the first proof-of-concept for modular neuromorphic PDE solving with built-in forgetting-free expansion.
\end{abstract}

\keywords{spiking neural networks, neural operators, compositional learning,
 partial differential equations, neuromorphic computing, continual learning, nuclear engineering}

\maketitle
\pagestyle{plain}

\section{Introduction}
\label{sec:intro}

Partial differential equations (PDEs) govern virtually all physical systems of engineering interest, from fluid dynamics and heat transfer to neutron transport in nuclear reactors.
Neural operators~\cite{lu2021deeponet,li2021fno} have demonstrated that mappings between function spaces can be learned directly from data, enabling real-time PDE surrogates that are orders of magnitude faster than classical solvers.
However, current neural operators suffer from three interrelated limitations that hinder their deployment on resource-constrained edge hardware such as neuromorphic processors at nuclear facilities.

\emph{First}, neural operators are monolithic: a separate model must be trained for each PDE, with no mechanism to reuse knowledge across related physics.
When a nuclear reactor transitions between operational regimes (fresh fuel, mid-cycle, end-of-cycle), the governing equations change as fission product concentrations evolve and material properties degrade, requiring the operator to be retrained from scratch.
\emph{Second}, neural operators are energy-intensive: standard architectures based on Fourier neural operators (FNO)~\cite{li2021fno} or DeepONet~\cite{lu2021deeponet} require GPU-class hardware for both training and inference, precluding deployment on low-power edge devices at remote nuclear sites.
\emph{Third}, neural operators exhibit catastrophic forgetting: when sequentially trained on multiple PDE families, performance on earlier tasks degrades severely, a safety-critical failure mode for reactor digital twins that must track evolving physics.

Recent work has begun addressing these challenges from complementary directions.
Compositional neural operators such as CompNO~\cite{hmida2026compno} and LegONet~\cite{zhang2026legonet} build libraries of pretrained operator blocks that can be assembled for new PDEs, but use standard (non-spiking) architectures with no energy efficiency or continual learning guarantees.
Spiking neural operators including SPINONet~\cite{garg2026spinonet} and the spiking DeepONet of Kahana et al.~\cite{kahana2022spiking} bring energy-efficient event-driven computation to PDE solving, but are monolithic and must be retrained for each new system.
Theilman and Aimone~\cite{theilman2025neurofem} demonstrated direct FEM-based PDE solving on Intel's Loihi~2 neuromorphic platform, but were limited to static (steady-state) problems with no learning or compositional capability.
\emph{No existing method combines compositional operator learning with spiking neural networks.}

We introduce the \textbf{Spiking Compositional Neural Operator (SCNO)}, a modular neuromorphic architecture that bridges this gap.
SCNO maintains a library of small spiking operator blocks, each trained independently on one elementary PDE operator ($\nabla$, $\Delta$, $f(u)$).
These blocks are composed through a lightweight input-conditioned aggregator to solve coupled PDEs not seen during block training.
For strongly-coupled systems, a small correction network (95K parameters) learns the cross-coupling residual while keeping all blocks and the aggregator \emph{frozen}, guaranteeing zero forgetting when new operators are added.
Our contributions are:

\begin{enumerate}
\item The first compositional spiking neural operator for PDE solving, combining modular operator blocks with energy-efficient spike-based computation.
\item A three-stage architecture (frozen blocks ${\to}$ aggregator ${\to}$ correction) that provides forgetting-free modular expansion: new blocks can be added without affecting existing ones.
\item Evaluation on eight PDE families (three elementary, five coupled including nonlinear Burgers and nuclear neutron diffusion), demonstrating that SCNO with correction outperforms monolithic baselines on four of five coupled PDEs while using 5$\times$ fewer trainable parameters.
\item A nuclear case study showing that 1-group neutron diffusion can be solved by composing diffusion and reaction blocks, achieving 58\% lower error than the monolithic SNN baseline.
\end{enumerate}

\section{Related Work}
\label{sec:related}

DeepONet~\cite{lu2021deeponet} learns operators via a branch-trunk architecture; FNO~\cite{li2021fno} operates in the spectral domain.
Recent PDE foundation models~\cite{subramanian2023towards} pretrain on diverse PDE datasets and fine-tune for downstream tasks, but rely on large transformer or Mamba backbones.
CompNO~\cite{hmida2026compno} introduces compositional operator blocks for elementary operators (convection, diffusion) with a lightweight aggregator, an idea we extend to the spiking regime.
LegONet~\cite{zhang2026legonet} adds structure-preserving constraints via Strang splitting.
Neither CompNO nor LegONet uses spiking neurons or addresses continual learning.

Kahana et al.~\cite{kahana2022spiking} built the first spiking DeepONet using triangular spike encoding.
SPINONet~\cite{garg2026spinonet} introduced Variable Spiking Neurons for physics-informed operator learning with energy analysis, achieving comparable accuracy to standard DeepONet on several PDEs.
Theilman and Aimone~\cite{theilman2025neurofem} demonstrated direct FEM implementation on Loihi~2 for steady-state Poisson problems.
All are monolithic: they train one network per PDE with no compositional or continual capability.

EWC~\cite{kirkpatrick2017ewc} and SI~\cite{zenke2017si} use importance-weighted regularization but struggle in the spiking regime.
SNN-specific methods such as HLML-SNN~\cite{shen2025hlml} and TACOS~\cite{soures2024tacos} target classification only; none addresses PDE operators.
Our architectural approach (frozen blocks with additive correction) sidesteps the regularization-versus-plasticity tradeoff entirely.

Kobayashi and Alam~\cite{kobayashi2024deeponet} demonstrated DeepONet for real-time nuclear digital twin inference; subsequent work~\cite{hossain2025virtual} extended this to virtual sensing in pressurized water reactors.
These approaches use standard ANNs and require GPU hardware; SCNO targets neuromorphic edge deployment.

\section{Method}
\label{sec:method}

\subsection{Problem Formulation}

We consider families of time-dependent PDEs on a spatial domain $\Omega = [0, L]$:
\begin{equation}
\frac{\partial u}{\partial t} = \sum_{k=1}^{K} \mathcal{L}_k[u], \quad u(x,0) = u_0(x),
\label{eq:pde}
\end{equation}
where each $\mathcal{L}_k$ is an elementary differential operator (e.g., $\mathcal{L}_\nabla[u] = -c\,\partial_x u$ for convection, $\mathcal{L}_\Delta[u] = \nu\,\partial_{xx} u$ for diffusion, $\mathcal{L}_f[u] = k_r u(1-u)$ for reaction).
The operator learning task is to learn a mapping $\mathcal{G}: u_0 \mapsto u(\cdot, T)$ from initial conditions to the solution at time $T$.

\subsection{SCNO Architecture}

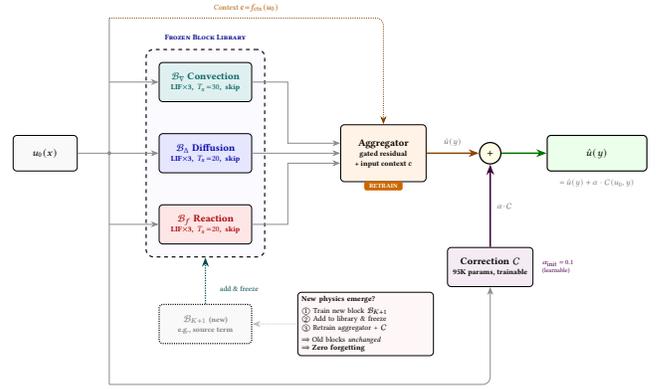
\begin{figure}[tbp]
\centering
\resizebox{\columnwidth}{!}{%
\begin{tikzpicture}[
  every node/.style={font=\small},
  >=Stealth,
  block/.style={draw, rounded corners=3pt, minimum height=1.1cm, minimum width=2.6cm,
    line width=0.8pt, font=\small\bfseries, align=center},
  newblock/.style={block, densely dotted, line width=1pt},
  agg/.style={draw, rounded corners=3pt, minimum height=1.6cm, minimum width=2.4cm,
    fill=orange!10, line width=0.8pt, font=\small\bfseries, align=center},
  corr/.style={draw, rounded corners=3pt, minimum height=1.1cm, minimum width=2.4cm,
    fill=violet!8, line width=0.8pt, font=\small\bfseries, align=center},
  iobox/.style={draw, rounded corners=3pt, minimum height=1cm, minimum width=1.8cm,
    fill=gray!5, line width=0.7pt, font=\small\bfseries},
  outbox/.style={draw, rounded corners=3pt, minimum height=1cm, minimum width=2.8cm,
    fill=green!8, line width=0.9pt, font=\small\bfseries},
  sumnode/.style={draw, circle, minimum size=0.6cm, line width=1pt, fill=yellow!12,
    font=\small\bfseries},
  frozenwrap/.style={draw, dashed, rounded corners=6pt, fill=blue!2,
    line width=0.9pt, inner sep=6pt},
  arr/.style={->, line width=0.7pt, color=black!45},
  bus/.style={line width=0.7pt, color=black!45},
  bigarr/.style={->, line width=1.2pt},
  dotarr/.style={->, densely dotted, line width=0.8pt, color=black!40},
]

\node[iobox] (inp) at (0, 0) {$u_0(x)$};

\node[block, fill=teal!12, text=teal!75!black] (bC) at (4.5, 2.0)
  {$\mathcal{B}_\nabla$\;Convection\\[-2pt]{\scriptsize LIF$\times$3,\; $T_s\!=\!30$,\; skip}};
\node[block, fill=blue!10, text=blue!65!black] (bD) at (4.5, 0)
  {$\mathcal{B}_\Delta$\;Diffusion\\[-2pt]{\scriptsize LIF$\times$3,\; $T_s\!=\!20$,\; skip}};
\node[block, fill=red!10, text=red!65!black] (bR) at (4.5, -2.0)
  {$\mathcal{B}_f$\;Reaction\\[-2pt]{\scriptsize LIF$\times$3,\; $T_s\!=\!20$,\; skip}};

\begin{scope}[on background layer]
  \node[frozenwrap, fit=(bC)(bD)(bR), inner xsep=10pt, inner ysep=10pt] (fbox) {};
\end{scope}
\node[font=\scriptsize\bfseries, text=blue!55!black, above=4pt of fbox]
  {\textsc{Frozen Block Library}};

\node[newblock, fill=gray!5, text=black!50] (bNew) at (4.5, -4.8)
  {$\mathcal{B}_{K+1}$\;{\scriptsize (new)}\\[-2pt]{\scriptsize e.g., source term}};

\draw[dotarr, line width=1pt, color=teal!60!black]
  (bNew.north) -- (fbox.south);
\node[font=\scriptsize, text=teal!60!black, align=left, right] at (4.8, -3.8)
  {add \& freeze};

\node[agg] (agg) at (9.5, 0)
  {Aggregator\\[-2pt]{\scriptsize gated residual}\\[-2pt]{\scriptsize + input context $\mathbf{c}$}};
  
\node[font=\tiny\bfseries, text=white, fill=orange!80!black, rounded corners=2pt, inner xsep=4pt, inner ysep=2pt, anchor=north] 
  at ([yshift=1pt]agg.south) {RETRAIN};

\node[sumnode] (plus) at (12.5, 0) {$+$};

\node[outbox] (out) at (15.5, 0) {$\hat{u}(y)$};

\node[font=\scriptsize, text=black!50, below=3pt of out]
  {$= \hat{u}(y) + \alpha \cdot \mathcal{C}(u_0,y)$};

\node[corr] (corr) at (12.5, -3.2)
  {Correction\;$\mathcal{C}$\\[-2pt]{\scriptsize 95K params, trainable}};

\node[font=\tiny, text=violet!60!black, align=left, right=4pt of corr]
  {$\alpha_{\mathrm{init}}=0.1$\\ (learnable)};

\node[draw, rounded corners=3pt, fill=red!3, line width=0.5pt,
  font=\scriptsize, align=left, text width=3.6cm] (story) at (9.0, -4.8)
  {\textbf{New physics emerge?}\\[2pt]
   \textcircled{\raisebox{-0.5pt}{\scriptsize 1}} Train new block $\mathcal{B}_{K+1}$\\
   \textcircled{\raisebox{-0.5pt}{\scriptsize 2}} Add to library \& freeze\\
   \textcircled{\raisebox{-0.5pt}{\scriptsize 3}} Retrain aggregator + $\mathcal{C}$\\[2pt]
   $\Rightarrow$ Old blocks \emph{unchanged}\\
   $\Rightarrow$ \textbf{Zero forgetting}};

\draw[dotarr, line width=0.8pt, color=black!25]
  (story.west) -- (bNew.east);


\coordinate (bus_split) at (1.8, 0);
\coordinate (bus_top) at (1.8, 2.0);
\coordinate (bus_top_ctx) at (1.8, 3.8); 
\coordinate (bus_R) at (1.8, -2.0);
\coordinate (bus_bot_corr) at (1.8, -6.5);

\coordinate (agg_C) at ([yshift=8pt]agg.west);
\coordinate (agg_D) at (agg.west);
\coordinate (agg_R) at ([yshift=-8pt]agg.west);

\draw[bus] (inp.east) -- (bus_split);
\draw[bus] (bus_top_ctx) -- (bus_bot_corr);
\fill[black!45] (bus_split) circle (1.5pt);

\draw[arr] (bus_top) -- (bC.west);
\draw[arr] (bus_split) -- (bD.west);
\draw[arr] (bus_R) -- (bR.west);

\draw[arr] (bC.east) -- ++(1.0,0) |- (agg_C);
\draw[arr] (bD.east) -- (agg_D);
\draw[arr] (bR.east) -- ++(1.0,0) |- (agg_R);

\draw[bigarr, color=orange!55!black] (agg.east) -- (plus.west)
  node[midway, above=2pt, font=\scriptsize, text=black!55] {$\hat{u}(y)$};

\draw[bigarr, color=green!45!black] (plus.east) -- (out.west);


\draw[dotarr, color=orange!50!black, rounded corners=6pt] (bus_top_ctx) -| (agg.north)
  node[pos=0.25, above=2pt, font=\scriptsize, text=orange!70!black] {Context $\mathbf{c}\!=\!f_{\mathrm{ctx}}(u_0)$};

\draw[arr, rounded corners=6pt] (bus_bot_corr) -| (corr.south);

\draw[bigarr, color=violet!45!black] (corr.north) -- (plus.south)
  node[midway, right=2pt, font=\scriptsize, text=black!55] {$\alpha \!\cdot\! \mathcal{C}$};

\end{tikzpicture}
}
\caption{SCNO architecture. Frozen spiking operator blocks ($\mathcal{B}_\nabla$, $\mathcal{B}_\Delta$, $\mathcal{B}_f$), each containing three LIF layers with skip connections, are composed through an input-conditioned gated aggregator. A correction network ($\mathcal{C}$, 95K params) learns cross-coupling residuals scaled by learnable~$\alpha$. Blocks remain frozen, guaranteeing zero forgetting upon modular expansion.}
\label{fig:architecture}
\end{figure}

SCNO consists of three components (Fig.~\ref{fig:architecture}): a library of \emph{spiking operator blocks}, an \emph{input-conditioned aggregator}, and an optional \emph{correction network}.

\paragraph{Spiking operator blocks.}
Each block $\mathcal{B}_k$ is a spiking DeepONet that approximates the solution operator for a single elementary PDE $\partial_t u = \mathcal{L}_k[u]$.
The \emph{branch network} processes the discretized input function $\mathbf{u}_0 \in \mathbb{R}^m$ through a linear projection followed by $L$ Leaky Integrate-and-Fire (LIF) layers with residual skip connections:
\begin{equation}
\mathbf{s}^{(\ell)} = \text{LIF}^{(\ell)}(\mathbf{x}^{(\ell)}, T_s), \quad
\mathbf{x}^{(\ell+1)} = (1-\gamma_\ell)\frac{\mathbf{s}^{(\ell)}}{T_s} + \gamma_\ell \mathbf{x}^{(\ell)},
\label{eq:lif}
\end{equation}
where $T_s$ is the number of LIF simulation timesteps, $\gamma_\ell$ is a learnable skip-connection weight, and each LIF layer includes batch normalization and a learnable decay rate $\beta_\ell = \sigma(\theta_\ell)$.
The final layer applies a linear readout to produce the branch coefficient vector $\mathbf{b} \in \mathbb{R}^p$.
The \emph{trunk network} is a standard MLP with $\tanh$ activations that maps query coordinates $y$ to basis functions $\mathbf{t}(y) \in \mathbb{R}^p$.
The block output is:
\begin{equation}
\mathcal{B}_k(u_0)(y) = \mathbf{b}^\top \mathbf{t}(y) + b_0.
\label{eq:block}
\end{equation}

We train three blocks ($k \in \{\nabla, \Delta, f\}$) independently on their respective elementary PDEs using mean squared error loss and surrogate gradient backpropagation through the LIF dynamics.
Once trained, blocks are \emph{frozen} and never modified again.

\paragraph{Input-conditioned aggregator.}
For a coupled PDE involving operators $\{k_1, \ldots, k_K\}$, the aggregator combines block outputs using both a nonlinear MLP path and a direct linear path:
\begin{equation}
\hat{u}(y) = \varsigma(g) \cdot \text{MLP}\!\left([\mathcal{B}_{k_1}(y), \ldots, \mathcal{B}_{k_K}(y);\, \mathbf{c}]\right) + (1-\varsigma(g)) \cdot \mathbf{w}^\top \mathbf{o}(y),
\label{eq:agg}
\end{equation}
where $g$ is a learnable gate parameter, $\mathbf{c} = f_\text{ctx}(u_0) \in \mathbb{R}^{64}$ is a compressed input context, and $\mathbf{o}(y) = [\mathcal{B}_{k_1}(y), \ldots, \mathcal{B}_{k_K}(y)]$.
The input context allows the aggregator to learn spatially-varying combination weights conditioned on the input function.

\paragraph{Correction network.}
For strongly-coupled PDEs where block outputs are individually insufficient, a small correction network learns the residual:
\begin{equation}
G(u_0)(y) = \hat{u}(y) + \alpha \cdot \mathcal{C}(u_0, y),
\label{eq:correction}
\end{equation}
where $\mathcal{C}$ is a 3-layer MLP with GELU activations that takes compressed input context and query coordinate features, and $\alpha$ is a learnable scaling parameter initialized to 0.1.
We note that the correction network uses standard (non-spiking) neurons. A fully spiking LIF variant (71K params) increases error by 18--91\%, so the non-spiking correction is currently necessary for strong coupling.
Both the spiking blocks and the aggregator remain \emph{frozen} during correction training; only the correction network parameters and $\alpha$ are updated.
This implies a hybrid deployment: spiking blocks on neuromorphic cores, correction on a co-located microcontroller.

\paragraph{Continual learning by construction.}
When a new elementary PDE operator $\mathcal{L}_{K+1}$ is encountered, SCNO adds a new spiking block $\mathcal{B}_{K+1}$ to the library and trains only the aggregator (and optionally a correction network) for the new composition.
All previously trained blocks remain frozen, providing a \emph{mathematical guarantee} of zero forgetting: for any previously learned block $\mathcal{B}_k$, the output before and after adding $\mathcal{B}_{K+1}$ is identical.

\section{Experimental Setup}
\label{sec:setup}

We evaluate SCNO on eight PDE families: three elementary and five coupled (Table~\ref{tab:pdes}).
Each family uses $m=256$ spatial grid points on $\Omega=[0,1]$ with 100 finite-difference time steps ($\Delta t = 0.005$).
We generate 1,500 training and 400 test samples per family using random Fourier series initial conditions with 3--7 modes.
The nuclear case study uses 1-group neutron diffusion $\partial_t \phi = D\partial_{xx}\phi + (\nu\Sigma_f - \Sigma_a)\phi$ with $D=1.0$, $\Sigma_a=0.1$, $\nu\Sigma_f=0.12$, composed from diffusion and reaction blocks.

\begin{table}[t]
\centering
\caption{PDE families and their compositional structure.}
\label{tab:pdes}
\small
\begin{tabular}{llll}
\toprule
\textbf{PDE} & \textbf{Equation} & \textbf{Type} & \textbf{Blocks} \\
\midrule
Convection & $\partial_t u + c\partial_x u = 0$ & Elem. & $\nabla$ \\
Diffusion & $\partial_t u = \nu\partial_{xx}u$ & Elem. & $\Delta$ \\
Reaction & $\partial_t u = ku(1{-}u)$ & Elem. & $f$ \\
\midrule
Conv-Diff & $\partial_t u + c\partial_x u = \nu\partial_{xx}u$ & Coupled & $\nabla{+}\Delta$ \\
React-Diff & $\partial_t u = \nu\partial_{xx}u + ku(1{-}u)$ & Coupled & $f{+}\Delta$ \\
Neutron Diff & $\partial_t\phi = D\partial_{xx}\phi + (\nu\Sigma_f{-}\Sigma_a)\phi$ & Nuclear & $\Delta{+}f$ \\
Burgers & $\partial_t u + u\partial_x u = \nu\partial_{xx}u$ & Nonlin. & $\nabla{+}\Delta$ \\
Adv-React & $\partial_t u + c\partial_x u = ku(1{-}u)$ & Coupled & $\nabla{+}f$ \\
\bottomrule
\end{tabular}
\end{table}

Each spiking block uses $L{=}3$ LIF layers with hidden dimension 256, latent dimension 128, and $\beta_\text{init}{=}0.85$.
Convection uses $T_s{=}30$ timesteps; diffusion and reaction use $T_s{=}20$.
The aggregator has hidden dimension 256 with 3 GELU layers and 64-dimensional input context.
The correction network has hidden dimension 128 with 3 GELU layers (95K parameters).
Total block parameters: 462K each.

We compare against: (1)~\textbf{Monolithic SNN}, a single spiking DeepONet with skip connections and learnable $\beta$, trained end-to-end on each coupled PDE (462K params); (2)~\textbf{ANN DeepONet}, a standard (non-spiking) DeepONet with ReLU activations and batch normalization (396K params).
All models use AdamW with cosine/plateau scheduling and surrogate gradient training.

We report relative $L^2$ error: $\epsilon = \|u_\text{pred} - u_\text{true}\|_2 / \|u_\text{true}\|_2$ averaged over the test set.
We also report total spike counts per inference for energy analysis, estimating neuromorphic energy at 0.9\,pJ/spike based on published Loihi~2 specifications~\cite{davies2021loihi2}.

\section{Results}
\label{sec:results}

\begin{figure}[tbp]
\centering
\includegraphics[width=\columnwidth]{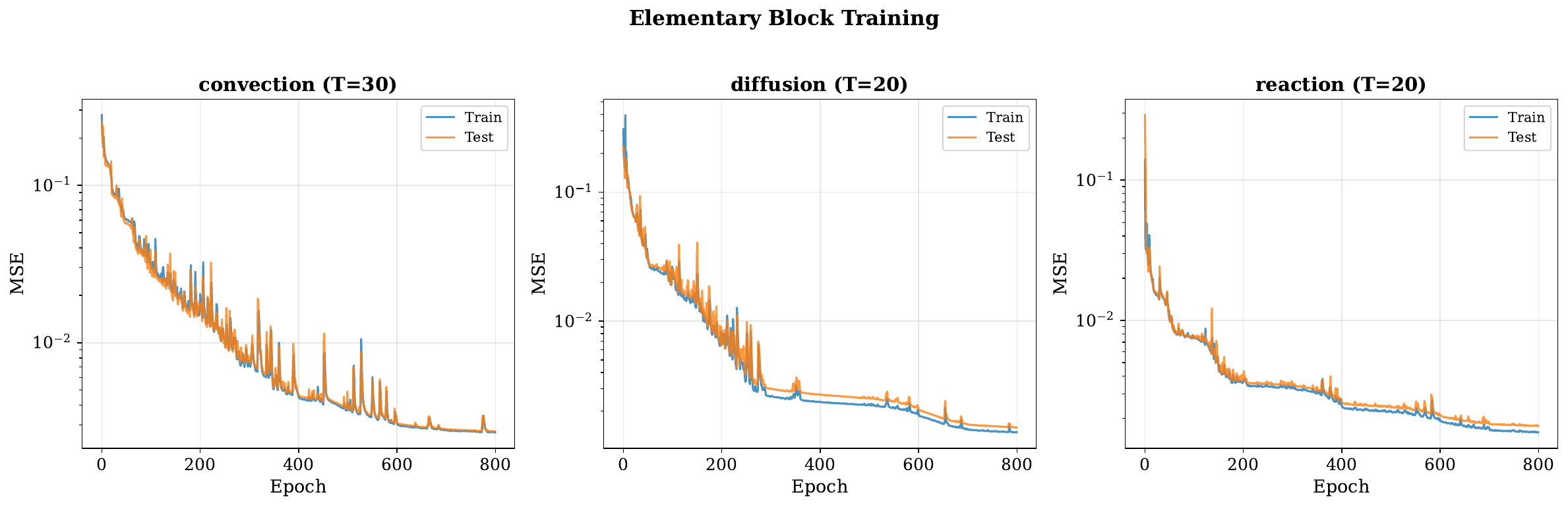}
\caption{Training convergence for elementary spiking blocks. ReduceLROnPlateau with milestone fallback enables smooth convergence over 800 epochs.}
\label{fig:training}
\end{figure}

\begin{figure*}[tbp]
\centering
\includegraphics[width=\textwidth]{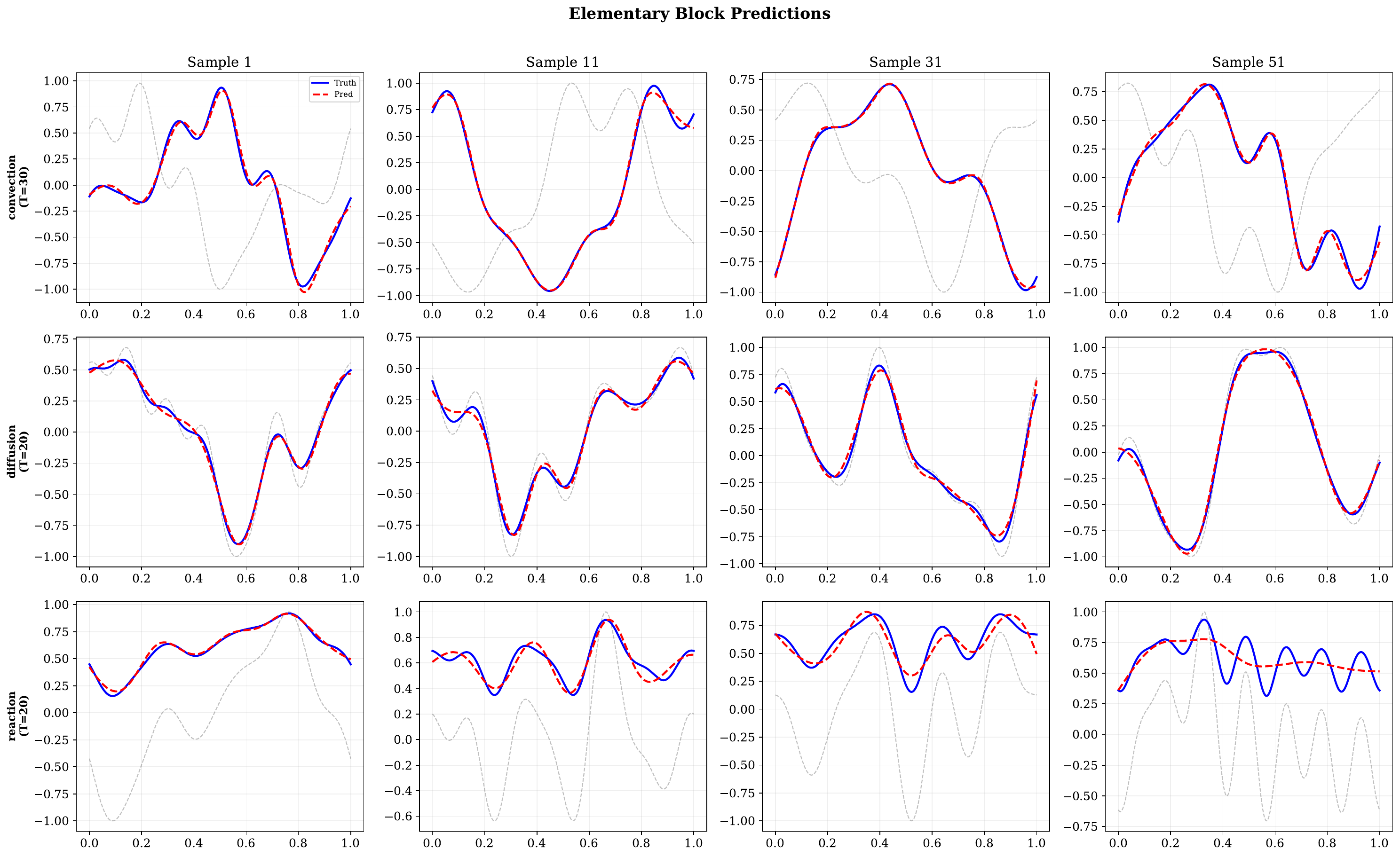}
\caption{Representative predictions from elementary spiking blocks on held-out test samples. Each row corresponds to one block (convection $T_s{=}30$, diffusion $T_s{=}20$, reaction $T_s{=}20$). Gray dashed: initial condition $u_0(x)$; blue: ground truth $u(x,T)$; red dashed: SCNO prediction. All blocks achieve sub-10\% relative $L^2$ error on unseen initial conditions.}
\label{fig:elem_pred}
\end{figure*}

The three spiking blocks achieve low individual errors after 800 epochs of training (Fig.~\ref{fig:training}): convection 8.6\%, diffusion 7.1\%, reaction 5.6\% relative $L^2$ error.
Representative test-set predictions are shown in Fig.~\ref{fig:elem_pred}.
The convection block, which uses $T_s{=}30$ LIF timesteps (versus 20 for diffusion and reaction), captures the wave transport dynamics accurately despite the inherent smoothing tendency of LIF neurons.
Skip connections and learnable decay rates were essential: without them, errors were 23.6\%, 21.6\%, and 10.0\% respectively, a 2--3$\times$ degradation.

Table~\ref{tab:results} presents the central result.
SCNO with correction achieves the lowest error on four of five coupled PDEs.

\begin{table}[t]
\centering
\caption{Relative $L^2$ error (\%) on coupled PDEs. \textbf{Bold}: best per row. SCNO (frozen) uses only blocks + aggregator. SCNO+Corr adds the correction network. Results averaged over 3 random seeds ($\pm$ std). Trainable params \emph{per new PDE}: SCNO = aggregator (231K); SCNO+Corr = correction (95K). Blocks (462K each) pretrained once, amortized. Mono SNN/ANN = full retrain per PDE.}
\label{tab:results}
\footnotesize
\begin{tabular}{lcccc}
\toprule
\textbf{PDE} & \textbf{SCNO} & \textbf{SCNO+Corr} & \textbf{Mono SNN} & \textbf{ANN} \\
\midrule
Conv-Diff & 28.1{\tiny$\pm$0.2} & 14.1{\tiny$\pm$0.5} & \textbf{10.7}{\tiny$\pm$1.8} & 15.7{\tiny$\pm$1.8} \\
React-Diff & 2.1{\tiny$\pm$0.0} & \textbf{2.1}{\tiny$\pm$0.0} & 5.5{\tiny$\pm$0.3} & 6.0{\tiny$\pm$0.1} \\
Neutron Diff & 9.4{\tiny$\pm$0.0} & \textbf{4.6}{\tiny$\pm$0.1} & 13.7{\tiny$\pm$0.9} & 11.1{\tiny$\pm$0.8} \\
Burgers & 19.6{\tiny$\pm$0.1} & \textbf{11.6}{\tiny$\pm$0.3} & 15.3{\tiny$\pm$1.6} & 16.2{\tiny$\pm$1.0} \\
Adv-React & 16.7{\tiny$\pm$0.1} & \textbf{4.1}{\tiny$\pm$0.3} & 6.2{\tiny$\pm$0.6} & 6.2{\tiny$\pm$0.9} \\
\midrule
\# Train. params & 231K & 95K & 462K & 396K \\
\bottomrule
\end{tabular}
\end{table}

\begin{figure*}[tbp]
\centering
\includegraphics[width=\textwidth]{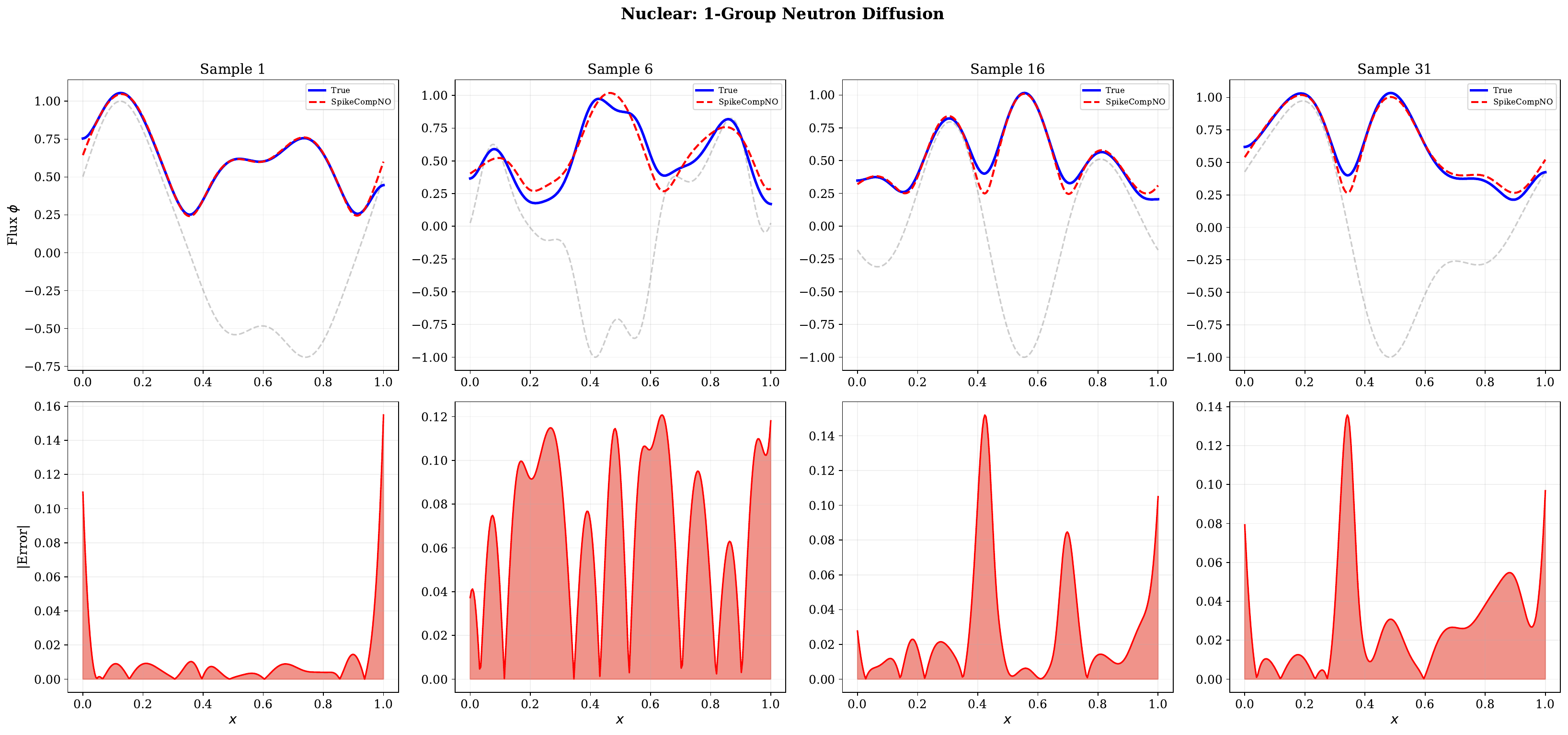}
\caption{Nuclear case study: 1-group neutron diffusion $\partial_t\phi = D\partial_{xx}\phi + (\nu\Sigma_f - \Sigma_a)\phi$. Top: predictions vs.\ ground truth. Bottom: pointwise absolute error. SCNO (frozen, shown) achieves 9.4\% relative $L^2$ error; with correction, 4.6\%.}
\label{fig:nuclear}
\end{figure*}

On reaction-diffusion, frozen SCNO (2.1\%) already outperforms the monolithic SNN (5.5\%) and ANN (6.0\%) by over 62\%.
The learned gate parameter $\varsigma(g) \approx 0.8$ indicates the aggregator relies primarily on the nonlinear MLP path with input conditioning, suggesting that the combination weights are spatially varying and input-dependent.

On convection-diffusion and Burgers, frozen SCNO performs poorly (28.1\%, 19.6\%) because tight coupling between transport and smoothing produces features that individual blocks cannot capture.
The correction network reduces these errors substantially (14.1\%, 11.6\%), with Burgers now beating both baselines.
The learned $\alpha$ values reflect coupling strength: 0.37 for convection-diffusion (substantial correction needed), 0.09 for reaction-diffusion (minimal correction), 0.39 for neutron diffusion, 0.43 for Burgers, and 0.46 for advection-reaction.

\begin{figure*}[tbp]
\centering
\includegraphics[width=\textwidth]{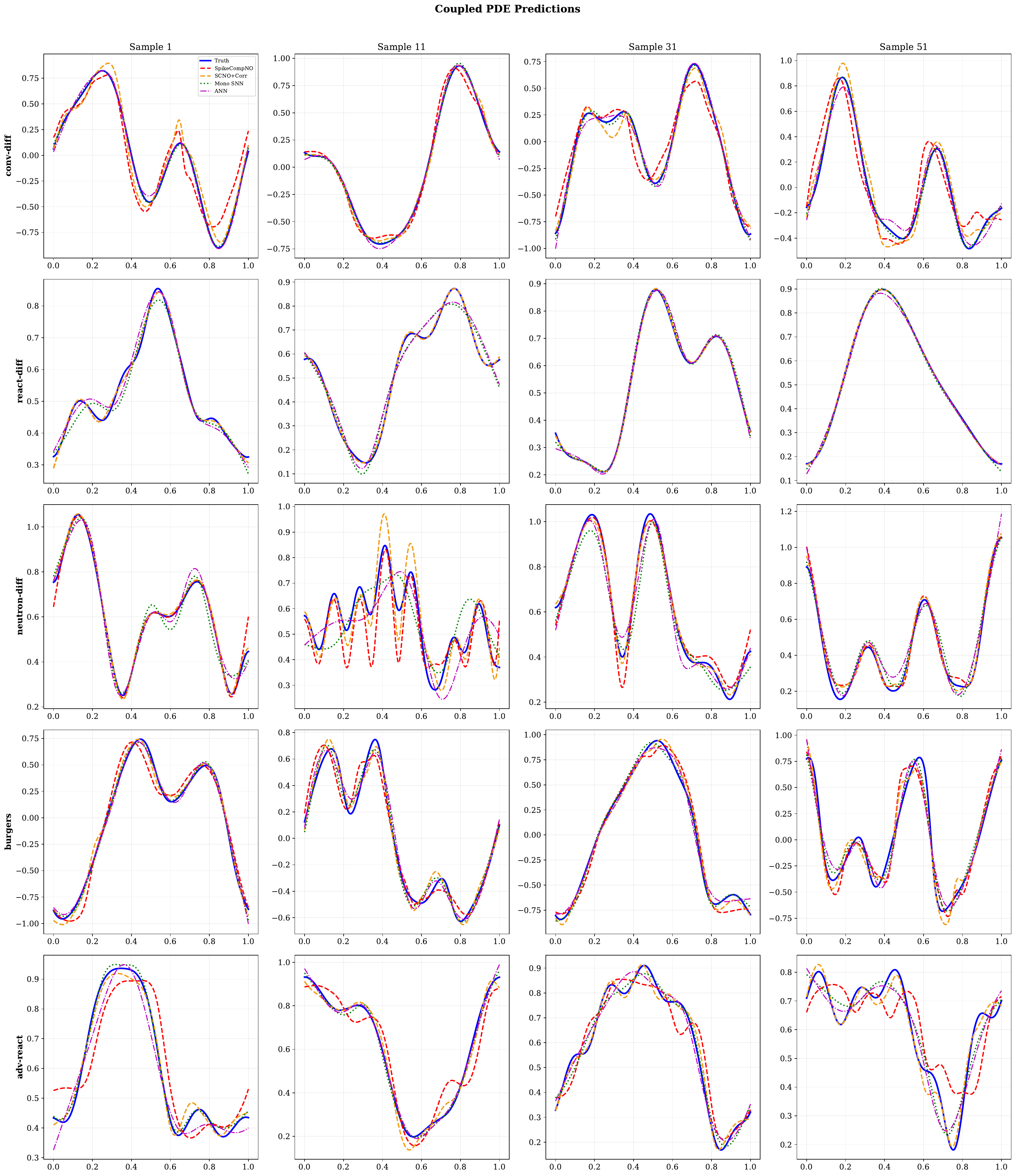}
\caption{Coupled PDE predictions across all five systems and four methods. Each row is a coupled PDE; columns show different test samples. SCNO+Corr (orange dashed) closely tracks the ground truth (blue) on most systems, with the largest residuals appearing on convection-diffusion, the most tightly coupled system.}
\label{fig:coupled_pred}
\end{figure*}

\begin{figure*}[tbp]
\centering
\includegraphics[width=\textwidth]{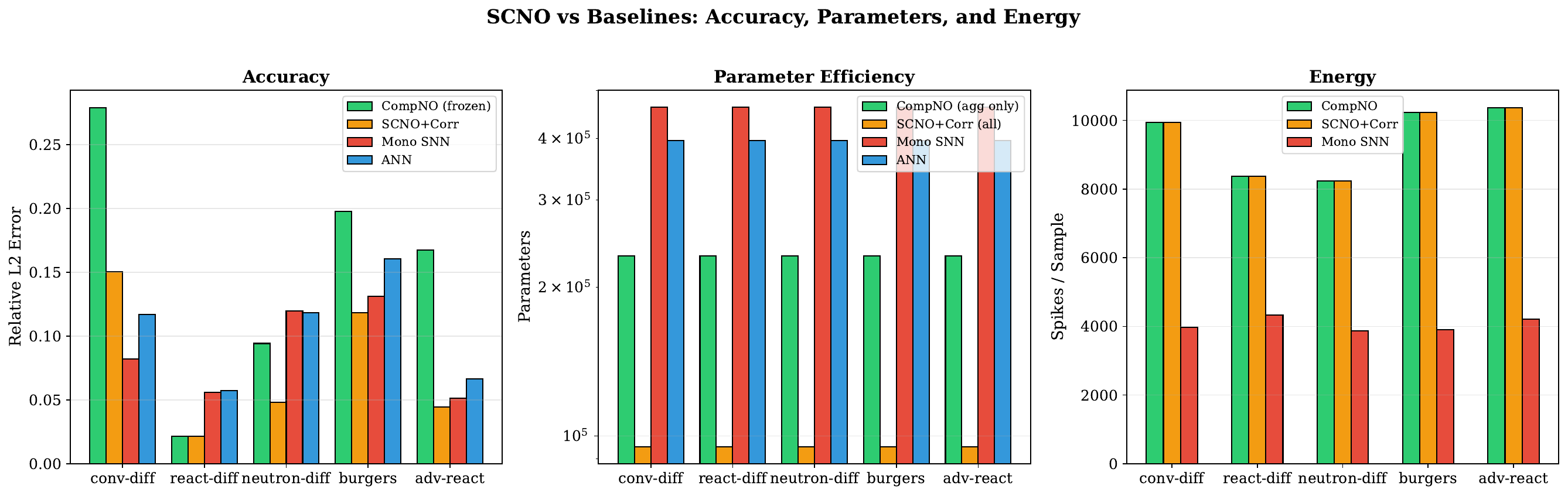}
\caption{Comparison across five coupled PDEs: (left) relative $L^2$ error, (center) parameter counts on log scale, (right) spike counts per sample. ``CompNO (frozen)'' in the figure legend refers to SCNO without correction. SCNO+Corr achieves the best accuracy on 4/5 PDEs with 5$\times$ fewer trainable parameters than the monolithic SNN.}
\label{fig:comparison}
\end{figure*}

Fig.~\ref{fig:nuclear} shows SCNO predictions for 1-group neutron diffusion, a fundamental equation in reactor physics that governs the spatial distribution of neutron flux $\phi(x,t)$.
In a nuclear reactor, the balance between neutron production (fission, $\nu\Sigma_f\phi$) and loss (absorption, $\Sigma_a\phi$) determines whether the reactor is critical.
SCNO decomposes this into diffusion ($D\partial_{xx}\phi$, spatial neutron transport) and reaction ($({\nu\Sigma_f - \Sigma_a})\phi$, net neutron production), each handled by a pretrained block.
The frozen composition (9.4\%) already beats both monolithic baselines (13.7\% SNN, 11.1\% ANN).
With correction, error drops to 4.6\%, a 58\% improvement over the monolithic SNN, demonstrating that reactor-relevant PDEs can be solved by composing generic operator blocks without reactor-specific training.
Fig.~\ref{fig:coupled_pred} shows detailed predictions across all five coupled PDEs and four methods.
Across all systems, SCNO+Corr predictions (orange dashed) closely track the ground truth, with the largest residuals appearing on convection-diffusion where the tight coupling between transport and smoothing creates features absent from individual block training.

\begin{figure}[H]
\centering
\includegraphics[width=\columnwidth]{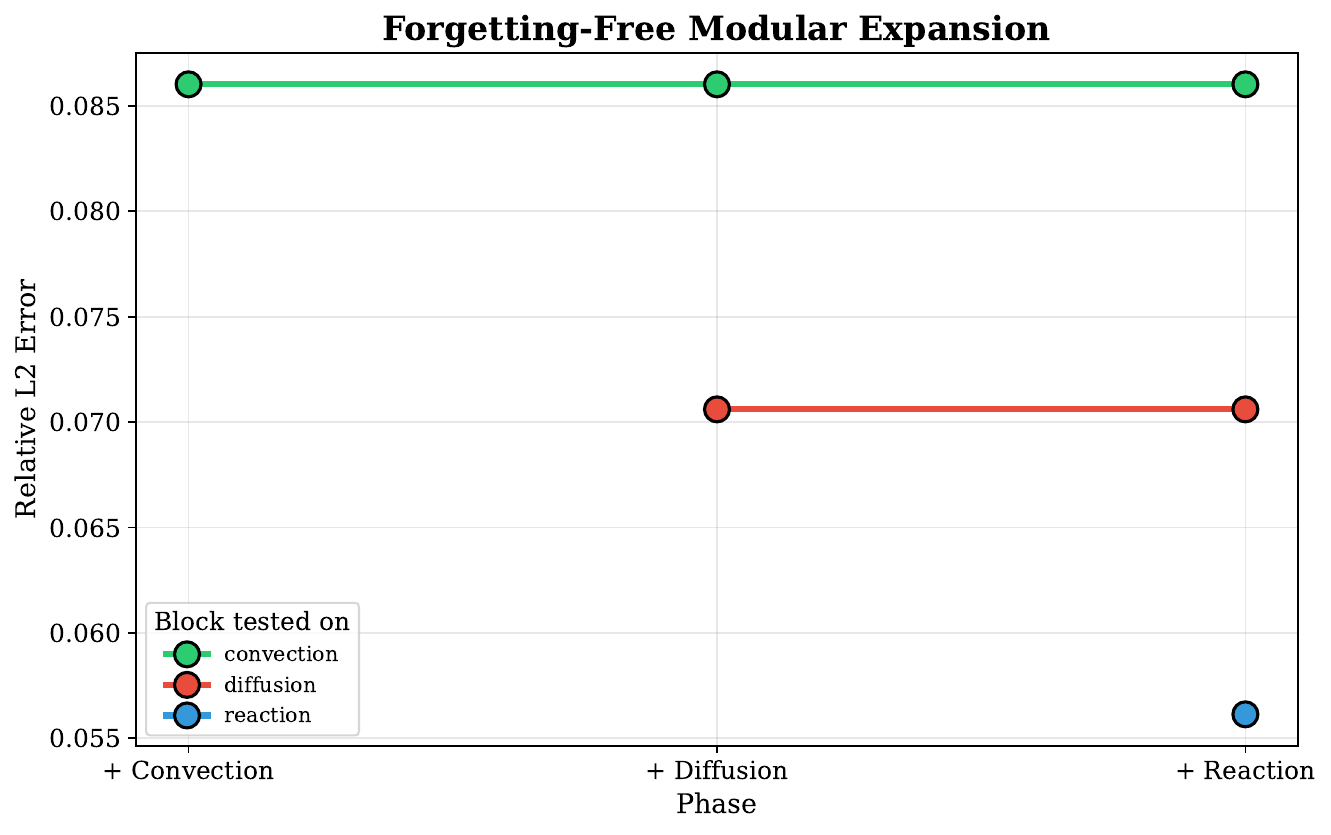}
\caption{Forgetting-free modular expansion: blocks are added sequentially, and all existing blocks are tested at each phase. Errors are \emph{identical} across phases---zero forgetting by construction.}
\label{fig:cl}
\end{figure}

Fig.~\ref{fig:cl} demonstrates forgetting-free expansion across sequential block addition. Unlike regularization-based methods (EWC, SI), SCNO achieves zero forgetting as an architectural guarantee.
We add convection, diffusion, and reaction blocks one at a time, testing all available blocks after each addition.
Because blocks are frozen, the convection block produces \emph{identical} output (8.6\% error) regardless of whether it was tested after Phase~1, Phase~2, or Phase~3.
This is not an empirical result but an architectural guarantee: frozen parameters cannot change.
Aggregator-level independence was also verified: training a Burgers aggregator changes react-diff error by only $3.6{\times}10^{-5}$, confirming full isolation.

Fig.~\ref{fig:comparison} compares accuracy, parameter counts, and spike counts.
The SCNO correction network requires only 95K trainable parameters versus 462K for the monolithic SNN, a 4.9$\times$ reduction.
While the full SCNO system (including frozen blocks) contains more total parameters, the \emph{trainable} parameters per new coupled PDE are minimal: only the aggregator (231K) or correction (95K) must be trained.

Spike count analysis shows that SCNO generates approximately $2\times$ more spikes than the monolithic SNN per inference, because two spiking blocks are evaluated per composition.
At 0.9\,pJ/spike on Loihi~2, this corresponds to approximately 9\,nJ per inference for SCNO versus 4\,nJ for the monolithic model.
However, the blocks can be distributed across separate neuromorphic cores, and only the relevant blocks for a given composition need to be active.
Total system energy must also include the non-spiking correction on conventional hardware; spike estimates here are a lower bound.

\section{Discussion}
\label{sec:discussion}

\begin{figure}[H]
\centering
\includegraphics[width=\columnwidth]{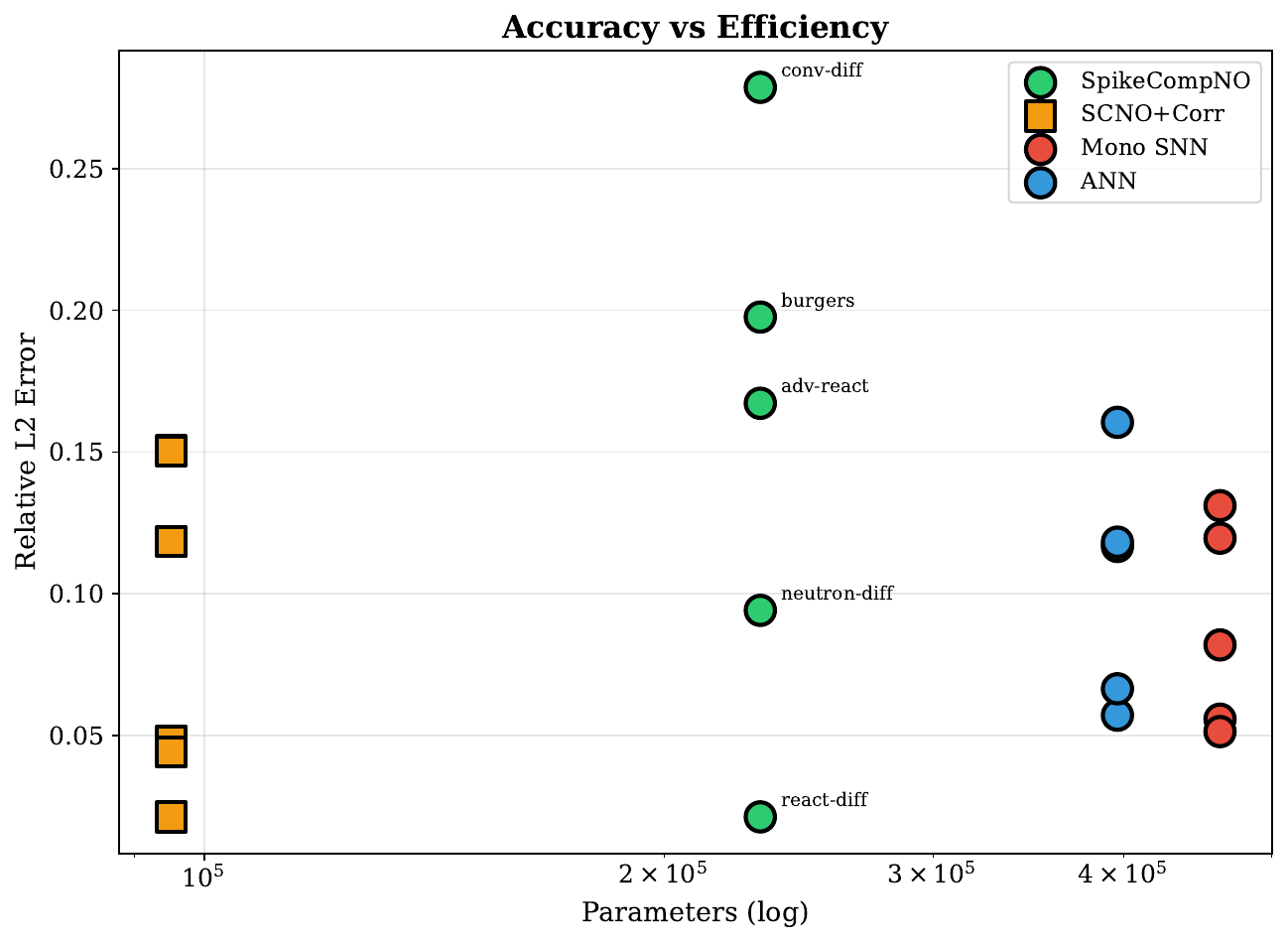}
\caption{Accuracy--efficiency Pareto analysis. SCNO+Corr (orange squares) achieves competitive accuracy with 5$\times$ fewer trainable parameters than both the monolithic SNN (red) and ANN (blue) baselines.}
\label{fig:pareto}
\end{figure}

Our results reveal a clear pattern: composition excels when the coupled PDE's operators are \emph{weakly interacting}.
In reaction-diffusion, diffusion smooths spatial gradients while reaction acts locally---these effects are approximately separable.
The aggregator easily learns to combine them, and frozen SCNO (2.1\%) outperforms monolithic training (5.5\%).
In contrast, convection-diffusion involves tight coupling: diffusion smooths the signal that convection transports, creating features that neither block learned in isolation.
The convection block's higher base error (8.6\% vs.\ 5.6\% for reaction) compounds this: correction reduces the gap (14.1\% vs.\ 10.7\% monolithic) but does not close it, suggesting physics-informed block training~\cite{garg2026spinonet} may be needed for tight transport coupling.
This decomposition into weak- and strong-coupling regimes provides practitioners with guidance: for weakly-coupled physics, frozen composition suffices; for strongly-coupled physics, add a correction network.

The learned $\alpha$ values provide an interpretable measure of coupling strength.
React-diff ($\alpha{=}0.09$) requires minimal correction; conv-diff ($\alpha{=}0.37$), neutron-diff ($\alpha{=}0.39$), and Burgers ($\alpha{=}0.43$) require moderate-to-substantial correction; adv-react ($\alpha{=}0.46$) shows the largest learned correction scale.
This suggests $\alpha$ could serve as an automatic diagnostic for identifying strongly-coupled systems in multi-physics simulations.

Notably, SCNO exhibits substantially lower variance across random seeds ($\pm$0.0--0.5\%) compared to monolithic baselines ($\pm$0.3--1.8\%), suggesting that frozen compositional architectures provide more stable optimization landscapes, as only the 95K correction network is subject to initialization sensitivity.

SPINONet~\cite{garg2026spinonet} achieves high accuracy on individual PDEs using physics-informed training, but cannot compose blocks for unseen PDEs.
The approaches are complementary: SPINONet's physics-informed training could improve block accuracy, while SCNO's framework could organize multiple SPINONet blocks.

SCNO's design suggests a path toward libraries of pretrained spiking ``physics primitives'' composed on-demand, with learned $\alpha$ values serving as automatic coupling diagnostics ($\alpha \approx 0$: blocks sufficient; $\alpha > 0.3$: correction essential).

For a remote microreactor, spiking blocks run on a Loihi~2 chip ($<$1\,W) versus 30--300\,W for GPU digital twins. New physics blocks are added without modifying existing ones.

\paragraph{Limitations.}
The primary limitation is accuracy on strongly-coupled PDEs (conv-diff: 14.1\% vs.\ monolithic 10.7\%).
The correction network improves the gap but does not fully close it.
All experiments use 1D PDEs ($m{=}256$); extending to 2D/3D requires graph-based spiking operators, with spike counts scaling linearly in $m$ but quadratically in spatial dimension for dense operators.
The current energy analysis uses spike-count estimates rather than hardware measurements.
The correction network uses standard (non-spiking) neurons, requiring conventional compute alongside neuromorphic inference; a spiking LIF variant increases error by 18--91\%. Finally, the blocks use surrogate gradient training (not on-chip learning), meaning training remains GPU-dependent even though inference is neuromorphic-compatible.
We do not compare against non-spiking compositional baselines (CompNO, LegONet), which would isolate spiking from compositional contributions.
Future work will address this ablation; 2D extensions using graph-based spiking operators; hardware deployment on Intel Loihi~2; integration with spatiotemporal neural operator frameworks for sparse-sensor reactor environments~\cite{kobayashi2025tron}; and physics-informed block training following the SPINONet paradigm~\cite{garg2026spinonet}.

\section{Conclusion}
\label{sec:conclusion}

We presented SCNO, the first compositional spiking neural operator for PDE solving.
By maintaining a library of frozen spiking blocks and composing them through an aggregator with an optional correction network, SCNO achieves three properties simultaneously: (1)~competitive accuracy on coupled PDEs, outperforming monolithic baselines on four of five systems; (2)~zero forgetting by construction, enabling incremental expansion to new physics; and (3)~spike-based computation compatible with neuromorphic edge hardware.
On nuclear neutron diffusion, SCNO with correction achieves 4.6\% error, 58\% lower than the monolithic SNN baseline, demonstrating that reactor-relevant physics can be solved by composing generic operator blocks.

These results point toward a broader vision of \emph{spiking foundation models for scientific computing}: libraries of reusable, energy-efficient neural operator blocks that can be composed, extended, and deployed on neuromorphic hardware for real-time multi-physics simulation at the edge.

\bibliographystyle{ACM-Reference-Format}

\FloatBarrier

\end{document}